\documentclass{article}

\usepackage{arxiv}

\usepackage[utf8]{inputenc} 
\usepackage[T1]{fontenc}    
\usepackage{hyperref}       
\usepackage{url}            
\usepackage{booktabs}       
\usepackage{amsfonts}       
\usepackage{nicefrac}       
\usepackage{microtype}      
\usepackage{lipsum}
\usepackage{graphicx}
\graphicspath{ {./images/} }

\usepackage{latexsym}
\usepackage{amsmath,amssymb,booktabs}
\usepackage{amsthm}
\usepackage{color}
\usepackage{wasysym}

\usepackage{bm}
\usepackage{braket}
\usepackage{multirow}
\usepackage{cite}
\usepackage{caption}
\usepackage{tabularx}
\usepackage[table]{xcolor}
\newcolumntype{C}{>{\centering\arraybackslash}X}
\newcommand{\graycell}{\cellcolor[gray]{0.92}}

\def\Underline{\setbox0\hbox\bgroup\let\\\endUnderline}
\def\endUnderline{\vphantom{y}\egroup\smash{\underline{\box0}}\\}
\def\|{\verb|}

\title{A Power-Weighted \\ Noncentral Complex Gaussian Distribution}

\author{
 Toru Nakashika \\
  Graduate School of Informatics and Engineering\\
  The University of Electro-Communications\\
  Tokyo 182-8585 \\
  \texttt{nakashika@uec.ac.jp}
}

\begin{document}
\maketitle
\begin{abstract}
The complex Gaussian distribution has been widely used as a fundamental spectral and noise model in signal processing and communication.
However, its Gaussian structure often limits its ability to represent the diverse amplitude characteristics observed in individual source signals.
On the other hand, many existing non-Gaussian amplitude distributions derived from hyperspherical models achieve good empirical fit due to their power-law structures, while they do not explicitly account for the complex-plane geometry inherent in complex-valued observations.
In this paper, we propose a new probabilistic model for complex-valued random variables, which can be interpreted as a power-weighted noncentral complex Gaussian distribution.
Unlike conventional hyperspherical amplitude models, the proposed model is formulated directly on the complex plane and preserves the geometric structure of complex-valued observations while retaining a higher-dimensional interpretation.
The model introduces a nonlinear phase diffusion through a single shape parameter, enabling continuous control of the distributional geometry from arc-shaped diffusion along the phase direction to concentration of probability mass toward the origin.
We formulate the proposed distribution and analyze the statistical properties of the induced amplitude distribution.
The derived amplitude and power distributions provide a unified framework encompassing several widely used distributions in signal modeling, including the Rice, Nakagami, and gamma distributions.
Experimental results on speech power spectra demonstrate that the proposed model consistently outperforms conventional distributions in terms of log-likelihood.
\end{abstract}


\section{Introduction}
Complex-valued data arise in many scientific and engineering fields, including speech and audio processing, signal processing, communications, and physics.
Typical examples include the spectrum obtained by the Fourier transform of speech signals, baseband signals in wireless communication, and complex representations of wave phenomena.
A fundamental probabilistic model for such complex-valued quantities is the complex Gaussian distribution, which has been widely used due to its mathematical tractability.

In speech and audio signal processing in particular, the circularly symmetric complex Gaussian distribution, represented by the Local Gaussian Model (LGM), has become a standard statistical model for signal spectra \cite{ephraim1984speech,ozerov2009multichannel,kitamura2016determined,fevotte2009nonnegative,ono2011stable,kameoka2009complex,sawada2013multichannel}.
While this model forms the basis of many signal processing methods, it relies on the strong assumption that amplitude and phase are statistically independent, making it difficult to adequately represent phase structures.
Although introducing a nonzero mean allows partial modeling of amplitude--phase dependence, the overall distributional shape remains Gaussian.
Moreover, the complex Gaussian distribution can be regarded as an approximation of the distribution obtained by the superposition of multiple source signals based on the central limit theorem.
Consequently, it is not necessarily suitable for representing the statistical characteristics of individual sources, such as heavy-tailed speech signals or light-tailed acoustic signals.

To alleviate some of these limitations, non-Gaussian, heavy-tailed distributions have been proposed.
For example, the complex generalized Gaussian distribution \cite{novey2009complex,mogami2020independent}, the complex Student-$t$ distribution \cite{mogami2017independent}, and the alpha-stable distribution \cite{leglaive2017alpha,mogami2022alpha} provide more flexible amplitude statistics.
However, these models assume circular symmetry with respect to the origin in the complex plane.
Under this assumption, the magnitude and phase components of a complex signal are considered statistically independent; in particular, the phase component is commonly simplified to follow a uniform distribution over the range 
$[0,\,2\pi)$.
In reality, the phase structure of short-time spectra in audio signals exhibits strong correlations due to the deterministic harmonic structure of the signals, rendering the uniform phase assumption inadequate for capturing physical reality.
Consequently, the geometric relationship between amplitude statistics and the complex-plane representation of the data has not been explicitly modeled.

Another important aspect in modeling speech spectral amplitudes is the statistical behavior on a logarithmic scale, which is closely related to human auditory perception.
Such logarithmic-scale characteristics are often modeled through power-law representations of amplitude or signal power.
To account for such characteristics, several amplitude distributions derived from higher-dimensional hyperspherical models have been studied, including the noncentral chi \cite{johnson1995continuous}, $\kappa$-$\mu$ \cite{yacoub2007kappamu,paris2014statistical}, and noncentral gamma distributions \cite{knusel1996computation}.
These models provide flexible descriptions of amplitude variability and have been widely used in signal and channel modeling.
However, these distributions implicitly assume higher-dimensional hyperspherical structures and do not explicitly incorporate the complex-plane geometry of complex-valued observations.
As a result, a mismatch arises between these models and the complex-valued observations.
Consequently, a probabilistic model that can simultaneously capture the coupling structure between amplitude and phase while consistently accounting for both linear- and logarithmic-scale characteristics has not yet been established.

In this paper, we propose a new probabilistic model for complex-valued random variables that addresses these limitations.
The proposed model can be interpreted as a power-weighted noncentral complex Gaussian distribution formulated directly on the complex domain.
Unlike conventional hyperspherical amplitude models, it consistently captures the geometric relationship between amplitude and phase while retaining the ability to represent diverse amplitude statistics.
The proposed model introduces a nonlinear phase diffusion on the complex plane through a single shape parameter, enabling continuous control of the distributional geometry from arc-shaped diffusion along the phase direction to concentration of probability mass toward the origin.
Note that this distribution is naturally derived from the energy function of the Boltzmann machine formulated in polar and log-polar coordinates \cite{nakashika2026logpolarbm}.
We systematically formulate the proposed distribution as a fundamental generative model for complex-valued random variables and theoretically analyze its statistical properties, including sampling methods and the derivation of $n$-th moments.

In particular, we show that the amplitude and power distributions derived from the proposed model constitute a highly unified framework that encompasses many well-known distributions as special cases, including the Rayleigh, Rice \cite{rice1945mathematical}, Nakagami \cite{nakagami1960m}, (one-sided) Gaussian, exponential, and gamma distributions, similarly to the noncentral gamma distribution derived from hyperspherical structures.
Furthermore, compared with the exponential distribution used as the power distribution in the LGM, its extension the gamma distribution, and even the noncentral gamma distribution, the proposed power distribution allows flexible control of tail heaviness through its parameters.
This property makes it suitable for representing a wide range of speech spectral characteristics, from sparse signals like voiced sounds to more complex acoustic signals containing consonant and reverberation.
Preliminary experimental results on speech spectra further support the effectiveness of the proposed model, showing consistent improvements over conventional power distributions.

The main contributions of this paper are summarized as follows:
\begin{enumerate}
\item We systematically formulate a generative model for complex-valued variables as a power-weighted noncentral complex Gaussian distribution.
\item We theoretically analyze the amplitude and power distributions and show that they can be interpreted as a unified framework encompassing several existing probability distributions.
\item We derive statistical quantities such as the mean, variance, higher-order moments, and kurtosis, thereby clarifying the statistical characteristics of the distribution.
\item We develop an efficient sampling method for generating random variables that follow the proposed distribution.
\item We empirically demonstrate that the proposed model consistently outperforms conventional power distributions in modeling speech power spectra.
\end{enumerate}

\begin{figure*}[tb]
    \centering
    \includegraphics[width=\linewidth]{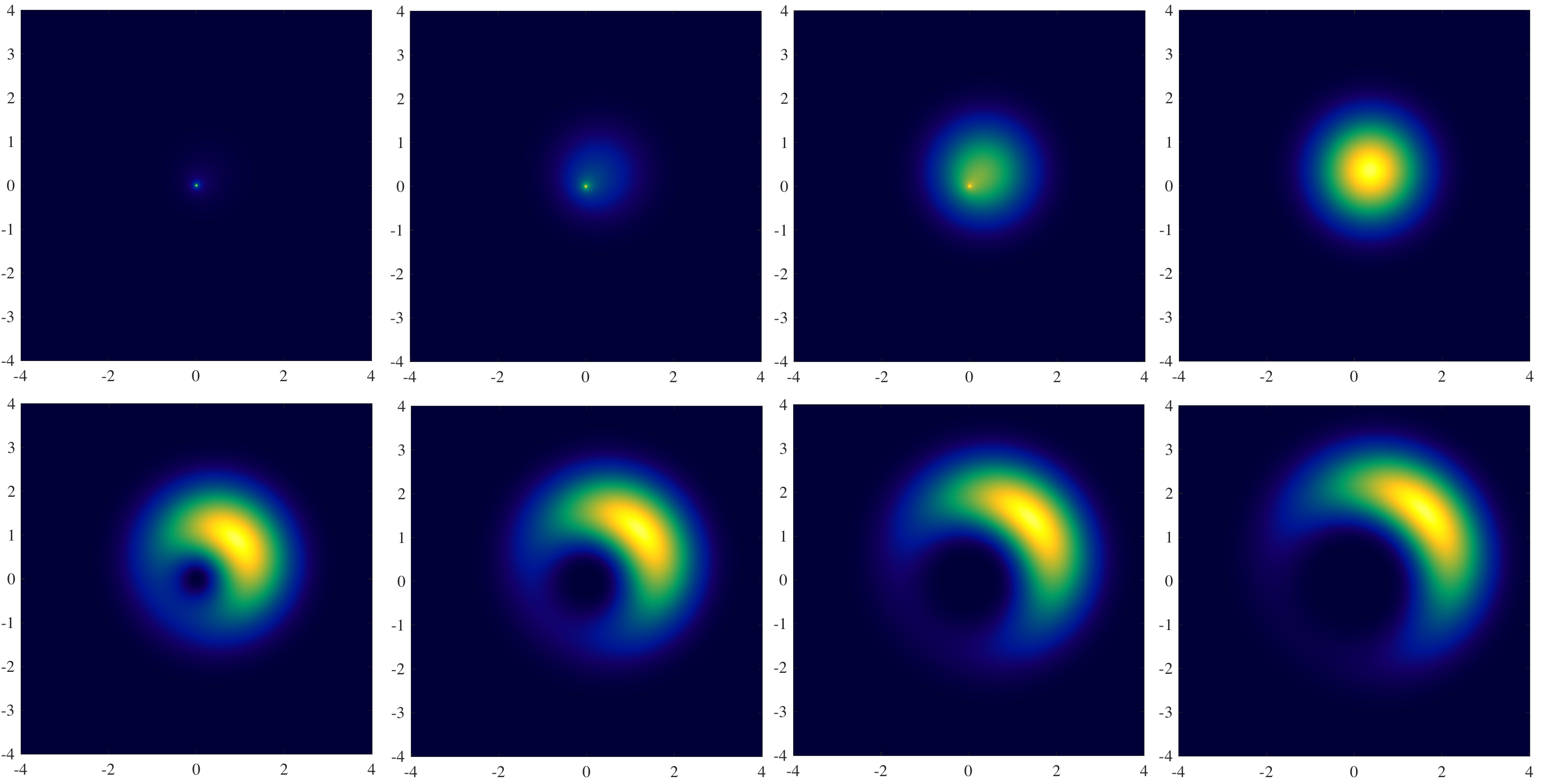}
    \caption{Effect of the shape parameter $\alpha$ on the proposed power-weighted noncentral complex Gaussian distribution. 
    From left to right in the top row: $\alpha=0.5, \, 0.75, \, 0.9, \, 1$; 
    from left to right in the bottom row: $\alpha=2, \, 3, \, 4, \, 5$. 
    The other parameters are fixed as $\sigma^2=1$ and $\mu=\frac{1}{2}{\rm e}^{\rm i \frac{\pi}{4}}$. 
    The horizontal and vertical axes represent the real and imaginary axes, respectively, 
    and lower and higher brightness indicate lower and higher probability density.}
    \label{fig:bcn_alpha}
\end{figure*}

\begin{figure*}[tb]
    \centering
    \includegraphics[width=\linewidth]{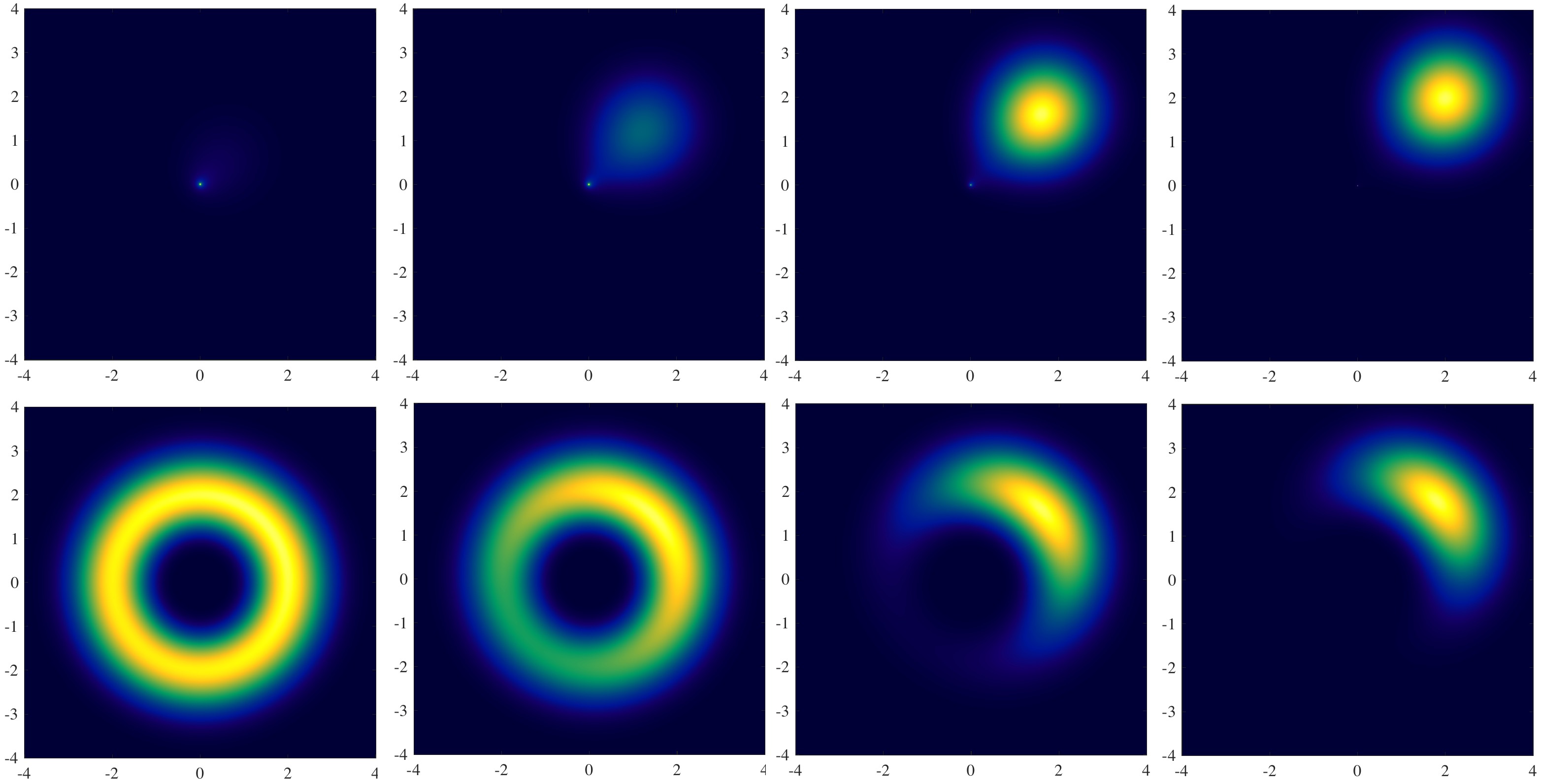}
    \caption{Effect of the amplitude of the mean parameter $|\mu|$ on the probability density of the proposed distribution. 
    Top row: $\alpha=0.5$ with $|\mu|=1, \, 2, \, 2.5, \, 3$ (from left to right). 
    Bottom row: $\alpha=5$ with $|\mu|=0.01, \, 0.1, \, 0.5, \, 1$ (from left to right). 
    The other parameters are fixed as $\sigma^2=1$ and $\angle \mu=\frac{\pi}{4}$. 
    The horizontal and vertical axes represent the real and imaginary axes, respectively, 
    and lower and higher brightness indicate lower and higher probability density.}
    \label{fig:bcn_mu}
\end{figure*}

\section{Proposed complex-valued distribution}

We define the proposed complex-valued probability density function by
\begin{align}
p(z; \mu, \sigma^2, \alpha) = \frac{|z|^{2\alpha-2} \mathrm{e}^{- \frac{|z-\mu|^2}{\sigma^2}}}{\pi \sigma^{2\alpha} \Gamma(\alpha)L_{\alpha-1}(-\frac{|\mu|^2}{\sigma^2})}, \label{eq:bcn}
\end{align}
where $z \in \mathbb{C}$ is a complex-valued variable, and $\mu \in \mathbb{C}$, $\sigma^2 \in \mathbb{R}_{>0}$, and $\alpha \in \mathbb{R}_{>0}$ denote the mean (centroid), variance, and shape parameter, respectively.
The functions $\Gamma(\cdot)$ and $L_\nu(\cdot)$ denote the gamma function and the Laguerre function of order $\nu \in \mathbb{R}$.
The proposed distribution can be interpreted as a power-weighted noncentral complex Gaussian distribution, where the radial power term $|z|^{2\alpha-2}$ modifies the density of a noncentral complex Gaussian distribution.
This additional weighting enables flexible control of the distributional geometry on the complex plane.
The function in \eqref{eq:bcn} satisfies the requirements of a probability density function, i.e.,
$p(z; \mu, \sigma^2, \alpha) \ge 0, \ \int_{\mathcal Z} p(z; \mu, \sigma^2, \alpha) dz = 1$,
(see Appendix~\ref{appendix:Z_R}).
As shown in Fig.~\ref{fig:bcn_alpha}, the parameter $\alpha$
significantly affects the shape of the distribution.
In particular, when $\alpha = 1$, the proposed distribution reduces to the conventional complex normal distribution:
\begin{align}
\mathcal{N}_{c}(z; \mu, \sigma^2) = \frac{\mathrm{e}^{- \frac{|z-\mu|^2}{\sigma^2}}}{\pi \sigma^{2}},
\end{align}
indicating that the proposed distribution can be regarded as an extension of the complex normal distribution.
For smaller values of $\alpha$ (i.e., $\alpha<1$), the probability density tends to concentrate near the origin, whereas for larger values of $\alpha$ (i.e., $\alpha>1$), the density near the origin decreases and nonlinear distortions appear along the phase direction.
This tendency becomes more pronounced when the magnitude of $\mu$ (i.e., $|\mu|$) is small, as illustrated in Fig.~\ref{fig:bcn_mu}.
In the top row of Fig.~\ref{fig:bcn_mu}, where $\alpha$ is fixed to a small value ($\alpha=0.5$), it can be observed that the convergence of probability density toward the origin becomes stronger as $|\mu|$ decreases.
Conversely, as shown in the bottom row of Fig.~\ref{fig:bcn_mu}, when $\alpha$ is fixed to a large value ($\alpha=5$), smaller values of $|\mu|$ lead to stronger diffusion along the phase direction.
Thus, based on the complex normal distribution, the proposed distribution allows control over its arc-shaped diffusion and convergence toward the origin through the newly introduced parameter $\alpha$.
In Figs.~\ref{fig:bcn_alpha} and \ref{fig:bcn_mu}, the mean phase is fixed to $\frac{\pi}{4}$ for visualization; varying the argument of $\mu$ simply rotates the distribution accordingly.
For sparse signals (e.g., voiced speech), smaller values of $\alpha$ increase the probability density near the origin, whereas larger values of $\alpha$ produce arc-shaped diffusion, which is suitable for non-sparse signals with relatively stable amplitudes (e.g., plosive sounds).
Hence, by adjusting $\alpha$, the proposed distribution can flexibly represent speech signals with diverse statistical characteristics.

Expressing Eq.~\eqref{eq:bcn} in polar coordinates, $z = r \mathrm{e}^{{\rm i} \theta}, \, r \in \mathbb{R}_{>0}, \, \theta \in [-\pi,\pi)$ yeilds the joint distribution of the amplitude $r$ and phase $\theta$.
Using the Jacobian $dz = r dr d\theta$, we obtain
\begin{align}
p(r,\theta) = \frac{r^{2\alpha-1} \mathrm{e}^{- \frac{r^2-2\nu r \cos(\theta-\phi) + \nu^2}{\sigma^2}}}{\pi \sigma^{2\alpha} \Gamma(\alpha)L_{\alpha-1}(-\frac{\nu^2}{\sigma^2})}, \label{eq:prtheta}
\end{align}
where we set $\mu = \nu \mathrm{e}^{{\rm i}\phi}, \, \nu \in \mathbb{R}_{\ge0}, \, \phi \in [-\pi,\pi)$.
Since Eq.~\eqref{eq:prtheta} contains the cross term $2\nu r \cos(\theta-\phi)$, the amplitude and phase are statistically dependent, and the joint distribution can be written as $p(r,\theta) = p(r)p(\theta|r)$.
The conditional distribution of the phase $p(\theta|r)$ is obtained by treating $r$ as a constant, yielding
\begin{align}
p(\theta|r) &\propto \mathrm{e}^{\frac{2 \nu r}{\sigma^2} \cos(\theta-\phi)} \\
&\propto \mathcal{VM}(\theta; \phi, \frac{2\nu r}{\sigma^2}),
\end{align}
which corresponds to a von Mises distribution $\mathcal{VM}$ with mean phase $\phi$ and concentration parameter $\frac{2\nu r}{\sigma^2}$.
On the other hand, the amplitude distribution $p(r)$ yields a new probability density function related to distributions such as the Rice and Nakagami distributions, as will be discussed in the next section.

\section{Derived amplitude distribution and related distributions}

\subsection{Proposed amplitude distribution}

We next discuss the amplitude distribution obtained by marginalizing the phase variable in Eq.~\eqref{eq:prtheta}.
The amplitude distribution is derived as
\begin{align}
p(r;\nu,\sigma^2,\alpha) &= \int_{-\pi}^{\pi} p(r,\theta) \, d\theta \\
&= \frac{r^{2\alpha-1} \mathrm{e}^{- \frac{r^2+\nu^2}{\sigma^2}}}{\pi \sigma^{2\alpha} \Gamma(\alpha) L_{\alpha-1}\!\left(-\frac{\nu^2}{\sigma^2}\right)}
\int_{-\pi}^{\pi} 
\mathrm{e}^{\frac{2 \nu r}{\sigma^2}\cos(\theta-\phi)} d\theta \\
&= \frac{r^{2\alpha-1} \mathrm{e}^{- \frac{r^2+\nu^2}{\sigma^2}}}{\pi \sigma^{2\alpha} \Gamma(\alpha) L_{\alpha-1}\!\left(-\frac{\nu^2}{\sigma^2}\right)}
\, 2\pi I_0\!\left(\frac{2\nu r}{\sigma^2}\right) \\
&= \frac{2r^{2\alpha-1} 
\mathrm{e}^{- \frac{r^2+\nu^2}{\sigma^2}}
I_0\!\left(\frac{2 \nu r}{\sigma^2}\right)}
{\sigma^{2\alpha} \Gamma(\alpha) L_{\alpha-1}\!\left(-\frac{\nu^2}{\sigma^2}\right)} ,
\label{eq:pr}
\end{align}
which yields a probability distribution similar to the Rice distribution.
Indeed, when $\alpha=1$, Eq.~\eqref{eq:pr} reduces to
\begin{align}
p(r;\nu,\sigma^2,1)
&= \frac{2r \mathrm{e}^{- \frac{r^2+\nu^2}{\sigma^2}}
I_0\!\left(\frac{2\nu r}{\sigma^2}\right)}
{\sigma^{2} \Gamma(1) L_0\!\left(-\frac{\nu^2}{\sigma^2}\right)} \\
&= \frac{2r}{\sigma^{2}}
\mathrm{e}^{- \frac{r^2+\nu^2}{\sigma^2}}
I_0\!\left(\frac{2\nu r}{\sigma^2}\right) \\
&= {\rm Rice}(r; \nu, \sigma^2),
\end{align}
which coincides with the Rice distribution.
Therefore, the proposed amplitude distribution can be regarded as a generalization of the Rice distribution.
Furthermore, in Eq.~\eqref{eq:pr}, by setting $\alpha=m$ and imposing the constraints
$\nu=0$ and $\sigma^2=\frac{\Omega}{m}$ with $\Omega>0$, we obtain
\begin{align}
p(r;0,\tfrac{\Omega}{m},m)
&= \frac{2 m^m r^{2m-1}
\mathrm{e}^{- \frac{m}{\Omega}r^2}}
{\Omega^{m} \Gamma(m)} \\
&= {\rm Nakagami}(r; m,\Omega),
\end{align}
which corresponds to the Nakagami-$m$ distribution.
Hence, the amplitude distribution in Eq.~\eqref{eq:pr} can also be interpreted as a generalization of the Nakagami distribution.
It also includes the Rayleigh distribution ($\nu=0$, $\alpha=1$) and the one-sided normal distribution ($\nu=0$, $\alpha=0.5$) as special cases.

\begin{figure}[tb]
    \centering
    \includegraphics[width=0.4\linewidth]{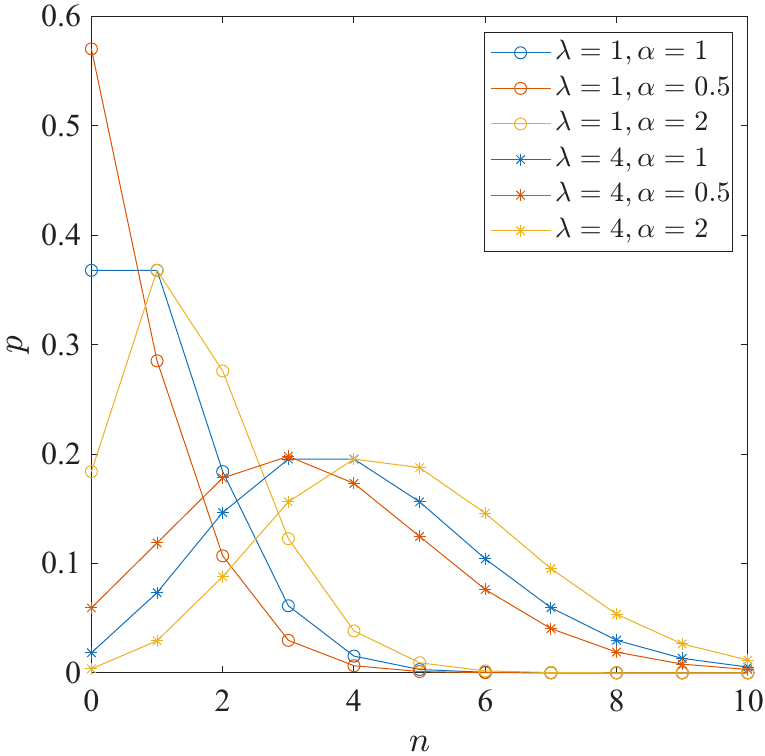}
    \caption{Examples of the Poisson-type distribution $p(n;\lambda,\alpha)$ distorted by the parameter $\alpha$.
    The blue curve corresponds to the standard Poisson distribution ($\alpha=1$),
    while the red and yellow curves represent the cases $\alpha<1$ and $\alpha>1$, respectively.}
    \label{fig:poisson_}
\end{figure}

Furthermore, consider the distribution obtained by setting the variance $\sigma^2=2$ and the shape parameter $\alpha=\frac{k}{2}$ with $k>0$:
\begin{align}
p(r;\nu,2,\frac{k}{2}) &= \frac{2^{1-\frac{k}{2}} r^{k-1} \mathrm{e}^{- \frac{r^2+\nu^2}{2}}I_0(\nu r) }{\Gamma(\frac{k}{2}) L_{\frac{k}{2}-1}(-\frac{\nu^2}{2})} \label{eq:chi_like}\\
&\propto r^{k-1} \mathrm{e}^{- \frac{r^2}{2}}I_0(\nu r).
\end{align}
This distribution resembles the noncentral chi distribution
\begin{align}
\chi' (r;k,\nu) &= \nu^{1-\frac{k}{2}} r^{\frac{k}{2}} \mathrm{e}^{- \frac{r^2+\nu^2}{2}}I_{\frac{k}{2}-1}(\nu r) \\
&\propto r^{\frac{k}{2}} \mathrm{e}^{- \frac{r^2}{2}}I_{\frac{k}{2}-1}(\nu r)
\end{align}
as well as the $\kappa$–$\mu$ distribution \cite{yacoub2007kappamu}.
In particular, when $k=2$, these distributions coincide, which reduces to the Rice distribution.
Therefore, the proposed distribution shares several properties with the noncentral chi distribution, such as describing the distribution of the distance of a complex-valued variable whose mean is shifted from the origin.
The key difference is that the order of the modified Bessel function in Eq.~\eqref{eq:pr} is always zero regardless of the power of $r$, whereas in the noncentral chi distribution the order of the modified Bessel function depends on the power of $r$.
This distinction clarifies the structural difference between the two distributions as follows.
Using a series expansion for the modified Bessel function
\[
I_{\frac{k}{2}-1}(x)
=
\sum_{n=0}^{\infty}
\frac{1}{n!\Gamma(n+\frac{k}{2})}
\left(\frac{x}{2}\right)^{2n+\frac{k}{2}-1},
\]
the noncentral chi distribution can be rewritten as
\begin{align}
\chi' (r;k,\nu) &= \sum_{n=0}^\infty \frac{{\rm e}^{-\frac{\nu^2}{2}}(\frac{\nu^2}{2})^n}{n!} \cdot \frac{r^{2n+k-1} {\rm e}^{-\frac{r^2}{2}}}{2^{n+\frac{k}{2}-1}\Gamma(n+\frac{k}{2})} \nonumber \\
&= \sum_{n=0}^\infty {\rm Poisson}(n; \frac{\nu^2}{2}) \chi(r;2n+k), \label{eq:chi_}
\end{align}
where ${\rm Poisson}(\cdot;\lambda)$ denotes the Poisson distribution and $\chi(\cdot;k)$ denotes the chi distribution.
As shown in Eq.~\eqref{eq:chi_}, the noncentral chi distribution can be interpreted as a mixture of chi distributions whose degrees of freedom are determined by a Poisson-distributed random variable $n$.
On the other hand, using the similar expansion
\[
I_0(x)
=
\sum_{n=0}^{\infty}
\frac{1}{n!\Gamma(n+1)}
\left(\frac{x}{2}\right)^{2n},
\]
the distribution in Eq.~\eqref{eq:chi_like} can be rewritten as
\begin{align}
p(r;\nu,2,\frac{k}{2}) &= \sum_{n=0}^\infty \frac{(\frac{k}{2})_n (\frac{\nu^2}{2})^n}{n!^2 L_{-\frac{k}{2}}(\frac{\nu^2}{2})} \cdot \frac{r^{2n+k-1} {\rm e}^{-\frac{r^2}{2}}}{2^{n+\frac{k}{2}-1}\Gamma(n+\frac{k}{2})} \nonumber \\
&= \sum_{n=0}^\infty p(n; \frac{\nu^2}{2}, \frac{k}{2}) \chi(r;2n+k), 
\end{align}
where $(\cdot)_n$ denotes the Pochhammer symbol and
\begin{align}
p(n;\lambda,\alpha)
=
\frac{(\alpha)_n \lambda^n}{n!^2 L_{-\alpha}(\lambda)}
\label{eq:poiss_}
\end{align}
defines a discrete distribution over nonnegative integers.\footnote{As shown in Appendix~\ref{appendix:poiss_proof}, this function satisfies the conditions of a probability distribution. Furthremore, this distribution actually coincides with the COM-NB distribution \cite{chakraborty2016poisson} with the power parameter fixed to two.}
When $\alpha=1$, we obtain
\begin{align}
p(n; \lambda, 1) &= \frac{(1)_n \lambda^n}{n!^2 L_{-1}(\lambda)} \\
&= \frac{n! \lambda^n}{n!^2 L_{0}(-\lambda) {\rm e}^\lambda} \\
&= \frac{{\rm e}^{-\lambda} \lambda^n}{n!} \\
&= {\rm Poisson}(n; \lambda),
\end{align}
which coincides with the Poisson distribution.
Therefore, the distribution in Eq.~\eqref{eq:poiss_} can be regarded as a generalization of the Poisson distribution.
Examples of this distribution are shown in Fig.~\ref{fig:poisson_}.
When $\alpha<1$, the distribution is skewed to the left relative to the Poisson distribution, whereas when $\alpha>1$, it is skewed to the right.
Consequently, both the noncentral chi distribution and the distribution in Eq.~\eqref{eq:pr}
can be interpreted as chi-mixture distributions in which the degrees of freedom are given by
$2n+k$ with a nonnegative integer $n$ following a certain discrete distribution.
The difference lies in the mixing distribution: the noncentral chi distribution employs the
Poisson distribution, whereas the proposed distribution uses its generalized form defined
in \eqref{eq:poiss_}.

\begin{figure}[tb]
    \centering
    \includegraphics[width=0.4\linewidth]{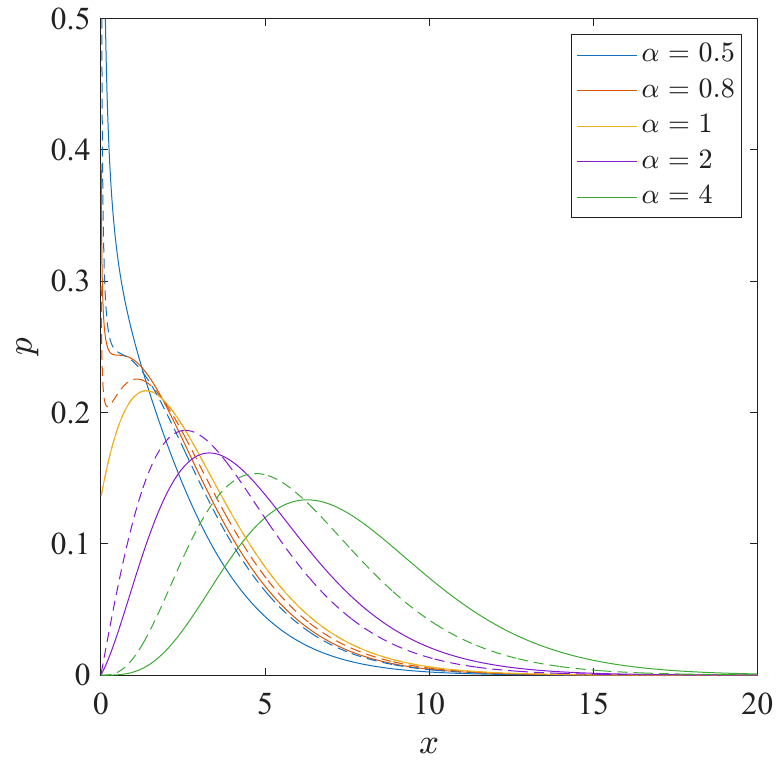}
    \caption{Proposed power distribution (solid lines) for different values of $\alpha$. The other parameters are fixed at $\beta=1$ and $\lambda=2$. The dashed lines represent the noncentral gamma distribution with the same parameters.}
    \label{fig:gamma_a}
\end{figure}

\begin{figure}[tb]
    \centering
    \includegraphics[width=0.4\linewidth]{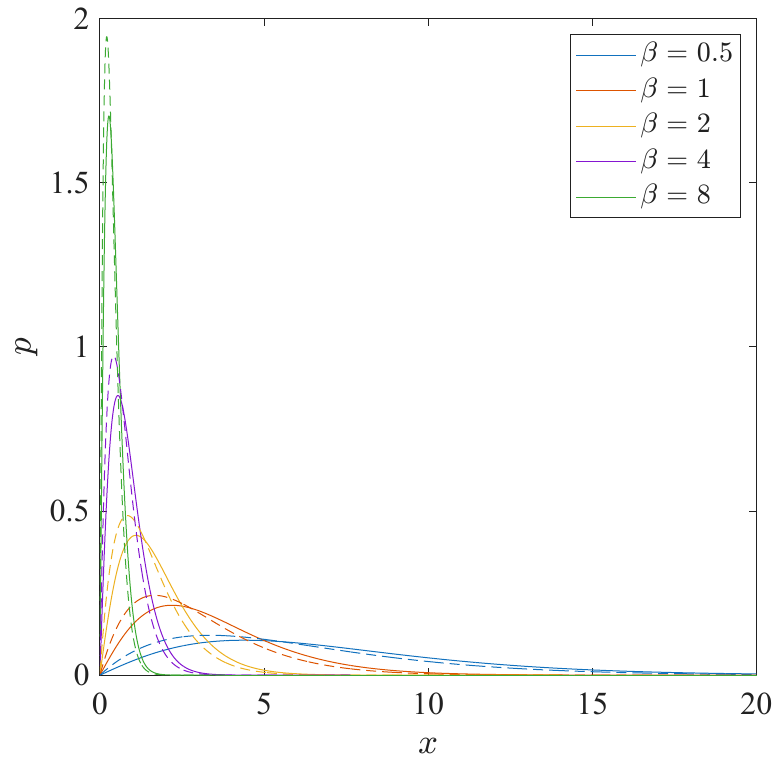}
    \caption{Proposed power distribution (solid lines) for different values of $\beta$. The other parameters are fixed at $\alpha=2$ and $\lambda=1$. The dashed lines represent the noncentral gamma distribution with the same parameters.}
    \label{fig:gamma_b}
\end{figure}

\begin{figure}[tb]
    \centering
    \includegraphics[width=0.4\linewidth]{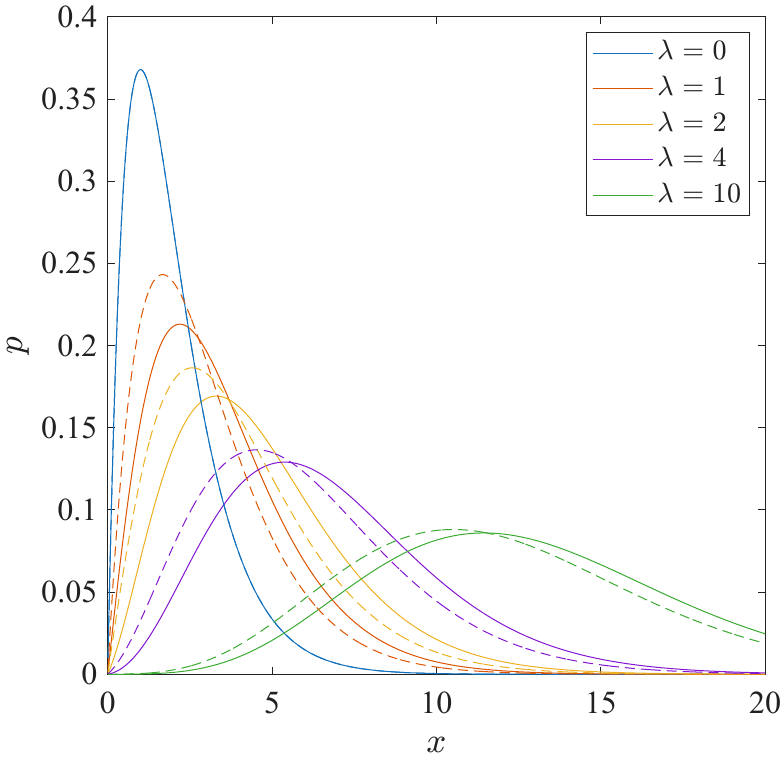}
    \caption{Proposed power distribution (solid lines) for different values of $\lambda$. The other parameters are fixed at $\alpha=2$ and $\beta=1$. The dashed lines represent the noncentral gamma distribution with the same parameters.}
    \label{fig:gamma_l}
\end{figure}

\subsection{Proposed power distribution}

The amplitude distribution in Eq.~\eqref{eq:pr} can be equivalently rewritten as a power distribution by applying the change of variables $x=r^2$.
Using the shape (degree) parameter $\alpha \in \mathbb{R}_{>0}$, the inverse variance (rate) parameter $\beta=\frac{1}{\sigma^2} \in \mathbb{R}_{>0}$, and the noncentrality parameter $\lambda = \frac{\nu^2}{\sigma^2} = \beta |\mu|^2 \in \mathbb{R}_{\ge 0}$, the distribution can be reformulated as
\begin{align}
p(x;\alpha,\beta,\lambda) &= \frac{\beta^{\alpha} }{\Gamma(\alpha) L_{-\alpha}(\lambda)} x^{\alpha-1} \mathrm{e}^{- \beta x}I_0(2\sqrt{\beta\lambda x}).
\label{eq:gammap}
\end{align}
This form is closely related to the noncentral gamma distribution, which is equivalent to the scaled noncentral chi-square distribution:
\begin{align}
&{\rm Gamma}'(x;\alpha,\beta,\lambda) = \frac{\beta^{\frac{\alpha+1}{2}} }{\lambda^{\frac{\alpha-1}{2}} \mathrm{e}^{\lambda}} x^{\frac{\alpha-1}{2}} \mathrm{e}^{- \beta x}I_{\alpha-1}(2\sqrt{\beta\lambda x}).
\end{align}
As discussed previously, the main difference is that the order of the modified Bessel function in the noncentral gamma distribution depends on the parameter $\alpha$, whereas in the proposed distribution it is always zero (the two distributions coincide when $\alpha=1$).
This difference reflects the fact that the noncentral gamma distribution represents the power of a $k=\frac{\alpha}{2}$-dimensional space, whereas the proposed power distribution represents the power obtained by projecting only the directional components of the high-dimensional space onto a two-dimensional complex space.

When $\lambda=0$, the distribution in Eq.~\eqref{eq:gammap} reduces to the gamma distribution
\begin{align}
{\rm Gamma}(x;\alpha,\beta) &= \frac{\beta^{\alpha} }{\Gamma(\alpha)} x^{\alpha-1} \mathrm{e}^{- \beta x}, \label{eq:gamma}
\end{align}
indicating that the proposed model can be regarded as a generalization of the gamma distribution (in this case it also coincides with the noncentral gamma distribution).
Since $\lambda=0$ corresponds to the case where the mean of the proposed complex distribution is located at the origin, the gamma distribution can also be interpreted as describing the power distribution of a complex variable with zero mean.
Furthermore, using the Poisson-type distribution defined in Eq.~\eqref{eq:poiss_}, Eq.~\eqref{eq:gammap} can also be written as
\begin{align}
p(x;\alpha,\beta,\lambda) = \sum_{n=0}^\infty p(n; \lambda, \alpha) {\rm Gamma}(x;n+\alpha,\beta). \label{eq:gammap2}
\end{align}
Therefore, the proposed power distribution can be interpreted as a mixture of gamma distributions with a Poisson distribution distorted by the degree parameter $\alpha$ as the mixing distribution, whereas the noncentral gamma distribution corresponds to a mixture of gamma distributions with the standard Poisson distribution.

Figs.~\ref{fig:gamma_a}, \ref{fig:gamma_b}, and \ref{fig:gamma_l} show examples of the proposed power distribution in Eq.~\eqref{eq:gammap} when the parameters $\alpha$, $\beta$, and $\lambda$ are varied, respectively.
For comparison, the noncentral gamma distribution with the same parameters is shown by dashed lines.
As shown in Fig.~\ref{fig:gamma_a}, the characteristics of the distribution differ significantly between $0<\alpha<1$ and $\alpha>1$: smaller values of $\alpha$ concentrate the probability density near the origin, whereas larger values reduce the density around the origin.
The parameter $\beta$ controls the scale of the distribution; smaller values make the distribution flatter and closer to a uniform distribution, whereas larger values cause the distribution to concentrate around its mean (Fig.~\ref{fig:gamma_b}).
In addition, as shown in Fig.~\ref{fig:gamma_l}, the noncentrality parameter $\lambda$ controls the deviation from the gamma distribution: when $\lambda$ is small the distribution approaches the gamma distribution, while larger values make it approach a one-sided normal distribution.

\subsection{Comparison with existing distributions}

In summary, as shown in Table~\ref{tab:pdfcomp}, the proposed distributions in Eqs.~\eqref{eq:pr} and \eqref{eq:gammap} simultaneously incorporate scale-equivariance (\textbf{Scale}), noncentrality (\textbf{N/C}), modeling in the complex domain (\textbf{Complex}), and higher-order structure (\textbf{Order}).
The proposed distribution can therefore be regarded as a higher-dimensional extension of the Rice distribution, which also models scale-equivariance and noncentrality in the complex domain.
Note that the proposed distribution differs from the noncentral gamma distribution, the noncentral chi distribution, and the $\kappa$-$\mu$ distribution, which represent amplitudes (or powers) on higher-dimensional hyperspheres.
While those distributions describe amplitudes arising from higher-dimensional spaces, the proposed distribution always models amplitudes (or powers) on a two-dimensional plane (i.e., the complex plane), regardless of the order parameter.
The Nakagami distribution represents the amplitude distribution of the sum of independent Gaussian variables in the complex plane and is therefore often interpreted as a complex-domain model.
Nevertheless, the distribution implicitly reflects a higher-dimensional structure determined by the number of Gaussian components.\footnote{Although the gamma distribution is essentially isomorphic to the Nakagami distribution and also represents a higher-dimensional structure, these distributions are distinguished in Table~\ref{tab:pdfcomp}. The Nakagami distribution was historically proposed to represent the amplitude distribution of the sum of independent Gaussian variables in the complex plane, where the number of components corresponds to the distribution parameter $m,(=\alpha)$. For this reason, it is regarded as modeling in the complex domain. In contrast, the gamma distribution does not assign such an interpretation to its parameter $\alpha$ and is instead a general statistical model for positive-valued variables unrelated to complex numbers.}
In contrast, the noncentral chi distribution, the noncentral gamma distribution, and the $\kappa$-$\mu$ distribution describe amplitude distributions on hyperspheres with nonzero centers.
The underlying directional distribution remains higher-dimensional, which can lead to inconsistencies when modeling complex-valued observations.
By contrast, the proposed distribution represents the amplitude distribution on a noncentral hypersphere whose directional component is projected onto the complex plane as a phase variable.
This property makes the model particularly suitable for complex-valued signals such as complex spectra.
The curvature observed in the phase direction in Figs.~\ref{fig:bcn_alpha} and \ref{fig:bcn_mu} can be interpreted as a consequence of projecting the directional components of a higher-dimensional hypersphere onto the phase of the complex plane.
While the amplitude distribution in Eq.~\eqref{eq:pr} and the power distribution in Eq.~\eqref{eq:gammap} are equivalent, the following discussion focuses on the latter due to its simpler form.

\begin{table}[t]
\centering
\caption{Comparison of the proposed distribution with conventional probability distributions.}
\label{tab:pdfcomp}
\begin{tabular}{l|cccc} \toprule
\textbf{Model} & \textbf{Scale} & \textbf{N/C} & \textbf{Complex} & \textbf{Order} \\ \midrule
\textbf{Rice} & \checkmark & \checkmark & \checkmark &   \\ 
\textbf{Nakagami} & \checkmark &       & \checkmark & \checkmark \\ 
$\chi'$ &          &  \checkmark  &   & \checkmark \\ 
$\kappa$-$\mu$ &  \checkmark &  \checkmark  &   & \checkmark \\ 
\textbf{Gamma} &  \checkmark &     &   & \checkmark \\ 
\textbf{Gamma}$'$ &  \checkmark &  \checkmark  &   & \checkmark \\ 
\textbf{Proposed} &  \checkmark & \checkmark &  \checkmark & \checkmark \\ \bottomrule
\end{tabular}
\end{table}

\section{Sampling from the proposed distribution}

Sampling is required in statistical inference based on Monte Carlo methods, such as the learning of Boltzmann machines \cite{nakashika2026logpolarbm}.  
In this section, we discuss a method for sampling complex random variables that follow the power-weighted noncentral complex Gaussian distribution defined in Eq.~\eqref{eq:bcn}.  
That is, we consider
\begin{align}
\tilde{z} \sim p(z; \mu, \sigma^2, \alpha).
\end{align}
Writing $z$ in polar form as $z = r {\rm e}^{\rm i \theta}$, where $r \in \mathbb{R}_{\ge 0}$ and $\theta \in [-\pi, \pi)$, the joint density factorizes as $p(r,\theta) = p(r)\,p(\theta|r)$.
Sampling $z$ can therefore be performed as
\begin{align}
\tilde{z} &= \tilde{r} {\rm e}^{\rm i \tilde{\theta}}, \\
\tilde{r} &\sim p(r), \\
\tilde{\theta} &\sim p(\theta|\tilde{r}) = \mathcal{VM}(\theta; \angle\mu, \frac{2|\mu|\tilde{r}}{\sigma^2}).
\end{align}
Sampling $r$ from the amplitude distribution $p(r)$ can be achieved by sampling the power variable $x$ using the power distribution in Eq.~\eqref{eq:gammap}:
\begin{align}
\tilde{r} &= \sqrt{\tilde{x}}, \\
\tilde{x} &\sim p(x; \alpha,\frac{1}{\sigma^2}, \frac{|\mu|^2}{\sigma^2}).
\end{align}
As shown in Eq.~\eqref{eq:gammap2}, the probability distribution of $x$ is a mixture of a Poisson-type distribution given in Eq.~\eqref{eq:poiss_} and a gamma distribution.  
Therefore, a sample $\tilde{x}$ can be obtained as
\begin{align}
\tilde{n} &\sim p(n; \frac{|\mu|^2}{\sigma^2}, \alpha), \\
\tilde{x} &\sim {\rm Gamma}(x; \tilde{n}+\alpha,\frac{1}{\sigma^2}).
\end{align}
Sampling from the gamma and von Mises distributions is straightforward since efficient sampling algorithms are well known.
Hence, sampling a complex random variable following the proposed complex distribution reduces to obtaining a sample $n$ from the Poisson-type distribution.
In this study, we propose two methods for generating random variables that follow the Poisson-type distribution $p(n; \lambda, \alpha)$: 
(i) a finite discrete distribution approximation method, and  
(ii) a Metropolis–Hastings sampling method.

\subsection{Sampling via finite discrete distribution approximation}

We first consider approximating the Poisson-type discrete distribution, whose normalization constant is given by an infinite series, by truncating the series at $N$ terms.
Specifically,
\begin{align}
p(n; \lambda, \alpha) &= \frac{\frac{\lambda^{n} (\alpha)_n}{n!^2}}{ \sum_{k=0}^\infty \frac{\lambda^{k} (\alpha)_k}{k!^2}} \\
&\approx \frac{\frac{\lambda^{n} (\alpha)_n}{n!^2}}{ \sum_{k=0}^N \frac{\lambda^{k} (\alpha)_k}{k!^2}}, \, \ n=0,1,\cdots, N.
\end{align}
Define the unnormalized term
\begin{align}
f(k)=\frac{\lambda^{k} (\alpha)_k}{k!^2},
\end{align}
which satisfies the recurrence
\begin{align}
f(0) &= 1 \\
f(k+1) &= \frac{\lambda (\alpha+k)}{(k+1)^2} f(k), \ k \ge 0.
\end{align}
Therefore, each term $f(k)$ and the cumulative sum $F(n) = \sum_{k=0}^n f(k)$ can be computed efficiently by iterative calculations.
Using the inverse transform sampling method, a random variable approximately following the Poisson-type distribution can be generated from a uniform random variable.
Specifically, given $u \sim {\rm U}(0,1)$, the sample $\tilde{n}$ is obtained by searching for the index satisfying
\begin{align}
F(\tilde{n}) < u \le F(\tilde{n}+1).
\end{align}
This method is simple and easy to implement.
However, it requires selecting an appropriate truncation number $N$.
If $N$ is too small, the approximation accuracy deteriorates.
On the other hand, choosing an unnecessarily large $N$ increases the computational cost proportionally.

\subsection{Metropolis--Hastings method}
\label{ssec:mh}

The Metropolis--Hastings method is a type of Markov chain Monte Carlo (MCMC) algorithm.
It allows sampling even when the normalization constant (here $L_{-\alpha}(\lambda)$) is difficult to compute, as in the case of the Poisson-type distribution.
In this study, we exploit the similarity between the Poisson-type distribution $p(n; \lambda, \alpha)$ and the standard Poisson distribution ${\rm Poisson}(n; \lambda)$, and use the Poisson distribution as both the initial distribution and the proposal distribution.
First, an initial candidate is drawn as $\tilde{n} \sim {\rm Poisson}(n; \lambda)$.
Next, a new candidate is generated from the proposal distribution $q(n) = {\rm Poisson}(n; \lambda)$, as $\tilde{n}' \sim q(n)$.
The acceptance probability $p$ is then computed as
\begin{align}
p &= \min\left(1, \, \frac{p(\tilde{n}'; \lambda,\alpha)}{p(\tilde{n}; \lambda,\alpha)}\cdot\frac{q(\tilde{n})}{q(\tilde{n}')} \right) \\
 &= \min \left(1, \,  \frac{\tilde{n}! \Gamma(\alpha+\tilde{n}')}{\tilde{n}'! \Gamma(\alpha+\tilde{n})} \right), \label{eq:mh_p}
\end{align}
whose derivation is given in Appendix~\ref{appendix:acceptrate}.
Notably, the acceptance ratio does not involve hypergeometric functions and can therefore be computed efficiently.
Based on this acceptance probability, the sample is updated using a uniform random variable $u \sim {\rm U}(0,1)$:
\begin{align}
\tilde{n} = \begin{cases}\tilde{n}' & (p>u), \\ \tilde{n} & ({\rm otherwize}). \end{cases}
\end{align}
The above steps (sampling a new candidate $\tilde{n}'$ and updating $\tilde{n}$) are repeated sufficiently many times.
Unlike the finite discrete distribution approximation, this method does not require selecting the truncation parameter $N$.
However, when $\alpha \gg 1$, the proposal distribution may significantly deviate from the Poisson-type distribution, which can degrade sampling efficiency.

\section{Statistical properties of the proposed power distribution}

In this section, we investigate the statistical properties of the power distribution defined in Eq.~\eqref{eq:gammap}.
The $n$-th moment $M_n$ of this distribution can be analytically computed as
\begin{align}
M_n = \mathbb{E}[x^n] &= \frac{(\alpha)_n}{\beta^{n}} \, \frac{L_{-\alpha-n}(\lambda)}{L_{-\alpha}(\lambda)}, \label{eq:mn}
\end{align}
(see Appendix~\ref{appendix:nth-moment} for the derivation).
Accordingly, the mean and variance are given by
\begin{align}
\mathbb{E}[x] &= \frac{\alpha}{\beta} \, R_\alpha(\lambda), \\
\mathbb{V}[x] &= \frac{\alpha}{\beta^2} \left( \lambda \frac{d R_\alpha(\lambda)}{d \lambda} + R_\alpha(\lambda) \right),
\end{align}
where
\begin{align}
R_\alpha(\lambda) &= \frac{L_{-\alpha-1}(\lambda)}{L_{-\alpha}(\lambda)}
\end{align}
denotes the ratio of Laguerre functions.
Furthermore, the moment generating function $M(t)$ can also be expressed in closed form as
\begin{align}
M(t) = \sum_{n=0}^\infty M_n \frac{t^n}{n!} = \left( \frac{\beta}{\beta-t} \right)^\alpha \frac{L_{-\alpha}(\frac{\beta \lambda}{\beta-t})}{L_{-\alpha}(\lambda)}.
\end{align}

\begin{figure}[tb]
    \centering
    \includegraphics[width=0.4\linewidth]{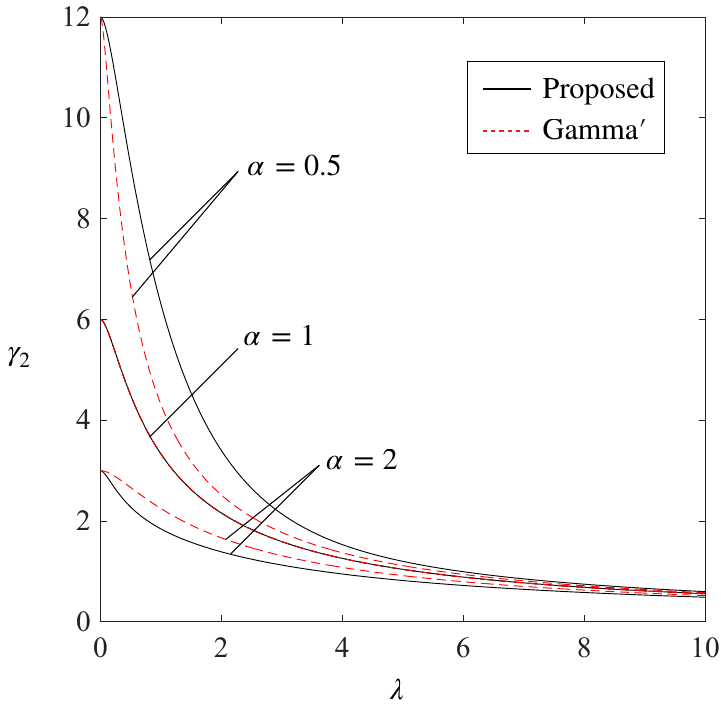}
    \caption{Comparison of the excess kurtosis of the proposed power distribution (black solid lines) and the noncentral gamma distribution (red lines) for varying $\lambda$ with $\alpha=0.5,\,1,\,2$.}
    \label{fig:kurt}
\end{figure}

Next, we examine the tail behavior and the concentration of probability mass around the origin.
The excess kurtosis $\gamma_2$ can be computed from the second and fourth cumulants $\kappa_2$ and $\kappa_4$, respectively, as
\begin{align}
\gamma_2 &= \frac{\kappa_4}{\kappa_2^2}, \\
\kappa_2 &= M_2 - M_1^2, \\
\kappa_4 &= M_4 - 4 M_1 M_3 - 3 M_2^2 + 12 M_1^2 M_2 - 6 M_1^4.
\end{align}
Figure~\ref{fig:kurt} shows the excess kurtosis $\gamma_2$ of the proposed power distribution and the noncentral gamma distribution\footnote{
The noncentral gamma distribution admits a closed-form expression for the $n$-th cumulant:
\begin{align}
\kappa_n = \frac{(n-1)!}{\beta^n} (\alpha + n \lambda).
\end{align}
}
for varying $\alpha$ and $\lambda$.
From the figure, it can be observed that the proposed power distribution spans a wider range of excess kurtosis than the noncentral gamma distribution.
Moreover, the relative behavior depends on $\alpha$: for $0 < \alpha < 1$, the proposed distribution exhibits larger excess kurtosis, whereas for $\alpha > 1$, the noncentral gamma distribution becomes more kurtotic, and both coincide at $\alpha=1$.
Note that the exponential distribution (corresponding to the central complex Gaussian model such as the LGM) is represented by the point $(\lambda,\gamma_2)=(0,6)$, the Rice distribution corresponds to the curve with $\alpha=1$, and the gamma distribution corresponds to a line on the $\gamma_2$-axis.
Therefore, both the noncentral gamma distribution and the proposed power distribution provide significantly more flexible modeling capability than these conventional distributions.
From a signal modeling perspective, impulsive components such as voiced speech or plosive sounds exhibit energy concentration at specific frequencies, resulting in sparse spectra and smaller values of $\alpha$.
In such cases, the heavier-tailed proposed power distribution can capture sharp spectral peaks while remaining robust to outliers, making it well suited for modeling such sparse signals.
On the other hand, more complex signals such as fricatives or reverberant mixtures tend to exhibit more diffuse energy distributions, approaching Gaussian behavior due to the central limit theorem.
For such non-sparse signals, the proposed distribution can also adapt by exhibiting lighter tails with larger $\alpha$.
These properties indicate that the proposed power distribution can theoretically model a wide range of acoustic signals with flexibility through appropriate parameter settings.

\begin{table*}[t]
\centering
\caption{Comparison of log-likelihood between the proposed power distribution and existing distributions: Individual scores, averages (Avg.), and $p$-values from one-tailed $t$-tests.}
\label{tab:res}
\begin{tabularx}{\textwidth}{l CCCCCCCCC} \toprule
\textbf{Model} & \textbf{FAKS0} & \textbf{MDAB0} & \textbf{FCMR0} & \textbf{MABW0} & \textbf{FCMH0} & \textbf{MBDG0} & \textbf{FAG0} & \textbf{MBNS0} & \textbf{FASW0} \\ \midrule
\textbf{Exp}      & 283.57 & 231.23 & 228.01 & 266.77 & 309.35 & 243.28 & 266.54 & 290.99 & 235.59 \\ 
\textbf{Gamma}    & 315.37 & 261.45 & 254.67 & 292.96 & 334.48 & 271.70 & 297.73 & 310.56 & 267.06 \\ 
\textbf{Gamma}$'$   & 315.63 & 261.66 & 254.95 & 293.16 & 334.71 & 271.85 & 297.90 & 310.71 & 267.41 \\ 
\textbf{Proposed} & \textbf{315.70} & \textbf{261.69} & \textbf{255.00} & \textbf{293.19} & \textbf{334.79} & \textbf{271.88} & \textbf{297.95} & \textbf{310.76} & \textbf{267.53} \\ 
\addlinespace[1ex]
\midrule
\textbf{Model} & \textbf{MAHH0} & \textbf{FDRW0} & \textbf{MCMJ0} & \textbf{FCAU0} & \textbf{MCHH0} & \textbf{FMLD0} & \textbf{MAJC0} & \graycell \textbf{Avg.} & \graycell \textbf{$p$-value} \\ \midrule
\textbf{Exp}      & 250.40 & 224.63 & 364.09 & 247.73 & 209.73 & 301.00 & 268.75 & \graycell 263.94 & \graycell $<.0001$ \\ 
\textbf{Gamma}    & 274.18 & 250.89 & 391.50 & 276.60 & 232.46 & 326.99 & 290.85 & \graycell 290.76 & \graycell $<.0001$ \\ 
\textbf{Gamma}$'$   & 274.38 & 251.22 & 391.73 & 276.78 & 232.64 & 327.16 & 291.07 & \graycell 290.98 & \graycell $<.0001$ \\ 
\textbf{Proposed} & \textbf{274.44} & \textbf{251.30} & \textbf{391.77} & \textbf{276.81} & \textbf{232.67} & \textbf{327.22} & \textbf{291.11} & \graycell \textbf{291.03} & \graycell - \\ \bottomrule
\end{tabularx}
\end{table*}

\section{Empirical verification}

To preliminarily validate the effectiveness of the proposed model, we conducted a fitting experiment on the power distribution of speech spectra.
We used the \texttt{TEST} set of the TIMIT speech corpus~\cite{timit}.
From each dialect region (\texttt{DR1} to \texttt{DR8}), one male and one female speaker were selected, resulting in a total of 16 speakers (e.g., \texttt{FAKS0}, \texttt{MDAB0}, \texttt{FCMR0}, \texttt{MABW0}, \texttt{FCMH0}, \texttt{MBDG0}, \texttt{FADG0}, \texttt{MBNS0}, \texttt{FASW0}, \texttt{MAHH0}, \texttt{FDRW0}, \texttt{MCMJ0}, \texttt{FCAU0}, \texttt{MCHH0}, \texttt{FMLD0}, \texttt{MAJC0}).
For each speaker, the utterance \texttt{SA1.WAV} was analyzed.
The short-time Fourier transform (STFT) was applied to each signal, and the squared magnitude of the resulting complex spectrum was used as the power.
The frame length and hop size were set to 16\,ms and 4\,ms, respectively.
If all time-frequency bins are pooled together, heterogeneous statistics arising from different frequencies and temporal variations are mixed.
To evaluate local statistical structures, we divided each spectrogram into small patches.
Specifically, the spectrogram was partitioned into regions of $3$ frequency bins and $20$ time frames, and the log-likelihood was computed for each region using the power values within the patch.
The final evaluation metric was obtained by averaging the log-likelihood over all patches.

Since direct evaluation of the complex-valued distribution defined in Eq.~\eqref{eq:bcn} is not straightforward, we instead compared the power distribution Eq.~\eqref{eq:gammap} with several conventional power distributions.
The compared models were the exponential distribution (\texttt{Exp}), the gamma distribution (\texttt{Gamma}), the noncentral gamma distribution (\texttt{Gamma}$'$), and the proposed power distribution (\texttt{Proposed}).
The parameters of each model were estimated by maximizing the log-likelihood.
For the exponential distribution, maximum likelihood estimation was performed in closed form, whereas for the gamma, noncentral gamma, and proposed distributions, parameters were optimized using a quasi-Newton method.

The obtained log-likelihoods for each speaker and their aggregated results are summarized in Table~\ref{tab:res}.
To assess the statistical significance of the improvement by the proposed method, paired one-sided $t$-tests were conducted.
The results show that, for all dialect regions and speakers, the proposed distribution consistently achieves significantly higher log-likelihoods than both the gamma and noncentral gamma distributions, demonstrating its superior flexibility in modeling power distributions.
In particular, substantial improvements over the exponential distribution, which is commonly used in the LGM framework, indicate that the proposed model can capture amplitude statistics that cannot be represented by simple exponential models.
On the other hand, the performance gap between the proposed model and the gamma or noncentral gamma distributions is relatively small.
This is attributed to the fact that the evaluation is performed solely on power (amplitude), where phase information is lost.
Indeed, the noncentrality parameter $\lambda$ of the noncentral gamma distribution frequently converges to zero, effectively reducing it to the standard gamma distribution.
This observation suggests that amplitude-only statistics are insufficient to capture asymmetry or directional structures inherent in complex-valued data.
Nevertheless, the proposed distribution consistently achieves the highest log-likelihood, indicating that its flexibility in shaping the distribution, particularly in controlling tail behavior, is effective.
It is worth noting that despite having a similar set of parameters, the proposed model significantly outperforms the noncentral gamma distribution, highlighting the advantage of directly modeling the geometric structure on the complex plane rather than relying on hyperspherical formulations.

Furthermore, since the proposed model is formulated directly on the complex domain, it can be naturally extended to incorporate phase structure, suggesting the potential for greater improvements in modeling full complex spectra.
Indeed, it was been demonstrated in \cite{nakashika2026logpolarbm} that the proposed complex-valued distribution also achieved strong performance in modeling speech complex spectra when combined with the Boltzmann machine framework.

\section{Conclusion}

In this paper, we proposed a novel complex-valued probabilistic model, referred to as the power-weighted noncentral complex Gaussian distribution, to overcome the limitations of conventional complex Gaussian models and hyperspherical models in signal processing.
The proposed model is formulated directly on the complex plane, enabling a consistent representation of the geometric relationship between amplitude and phase, while retaining the flexibility of higher-dimensional amplitude statistics.
The distribution introduces nonlinear phase diffusion controlled by a single parameter, allowing continuous interpolation between arc-shaped phase diffusion and concentration of probability mass toward the origin.
This enables the model to capture both sparse structures, such as voiced components, and non-sparse structures arising from signal superposition, thereby providing a precise representation of complex spectra with diverse amplitude characteristics.
Through theoretical analysis, we showed that the derived amplitude and power distributions form a highly general framework encompassing many well-known distributions, including the Rice, Nakagami, and gamma-type distributions.
Experimental validation on speech power spectra demonstrated that the proposed model consistently achieves higher log-likelihoods than conventional gamma-based distributions, supporting its effectiveness for flexible amplitude modeling.
Future work will focus on further exploring various acoustic tasks, such as speech synthesis and sound source separation, while extending the proposed model to a broader range of engineering problems, including medical image analysis and communication systems.

\section*{Acknowledgment}
This work was partially supported by JSPS KAKENHI Grant Number 24H00715. The author would also like to thank Prof.~Kohei Yatabe for his valuable comments.

\appendix

\section{Proof of $\int_\mathcal{Z} p(z; \mu, \sigma^2, \alpha)\, dz = 1$}
\label{appendix:Z_R}
\begin{proof}
We show that the power-weighted noncentral complex Gaussian distribution defined in Eq.~\eqref{eq:bcn} satisfies the normalization condition of a probability density function: $\int_\mathcal{Z} p(z; \mu, \sigma^2, \alpha) dz = 1$.
That is,
\begin{align}
1=&\int_\mathcal{Z} p(z; \mu, \sigma^2, \alpha) dz \\
=& \int_\mathcal{Z} \frac{|z|^{2\alpha-2} \mathrm{e}^{- \frac{|z-\mu|^2}{\sigma^2}}}{\pi \sigma^{2\alpha} \Gamma(\alpha) L_{\alpha-1}(-\frac{|\mu|^2}{\sigma^2})} dz \\
=& \frac{\mathrm{e}^{- \frac{|\mu|^2}{\sigma^2}}}{\pi \sigma^{2\alpha} \Gamma(\alpha) L_{\alpha-1}(-\frac{|\mu|^2}{\sigma^2})} \int_\mathcal{Z} |z|^{2\alpha-2} \mathrm{e}^{- \frac{|z|^2-2 \Re [\bar{\mu}z]}{\sigma^2}} dz.
\end{align}
Hence, it suffices to show that
\begin{align}
&\int_\mathcal{Z} |z|^{2\alpha-2} \mathrm{e}^{- \frac{|z|^2-2 \Re [\bar{\mu}z]}{\sigma^2}} dz \label{eq:a1} \\
&= \frac{\pi \sigma^{2\alpha} \Gamma(\alpha)L_{\alpha-1}(-\frac{|\mu|^2}{\sigma^2})}{\mathrm{e}^{- \frac{|\mu|^2}{\sigma^2}}}. \label{eq:a2}
\end{align}
We evaluate the left-hand side. By rewriting the exponent and transforming into polar coordinates $z = r e^{i\theta}$, we obtain
\begin{align}
&\int_\mathcal{Z} |z|^{2\alpha-2} \mathrm{e}^{- \frac{|z|^2-2 \Re [\bar{\mu}z]}{\sigma^2}} dz \\
=& \int_\mathcal{Z} |z|^{2\alpha-2} \mathrm{e}^{- \frac{|z|^2}{\sigma^2} + \frac{2|\mu z| \cos(\angle z-\angle \mu)}{\sigma^2}}dz \\
=& \int_0^\infty \int_{-\pi}^\pi r^{2\alpha-2} \mathrm{e}^{- \frac{r^2}{\sigma^2} + \frac{2|\mu|r}{\sigma^2} \cos(\theta-\angle \mu)} rdr d\theta \\
=& \int_0^\infty r^{2\alpha-1} \mathrm{e}^{- \frac{r^2}{\sigma^2}} \int_{-\pi}^\pi \mathrm{e}^{\frac{2|\mu|r}{\sigma^2} \cos(\theta-\angle \mu)}d\theta dr \\
\intertext{Using the integral representation of the modified Bessel function,}
=& \int_0^\infty r^{2\alpha-1} \mathrm{e}^{- \frac{r^2}{\sigma^2}} 2\pi I_0(\frac{2|\mu|r}{\sigma^2}) dr \\
\intertext{Expanding $I_0(\cdot)$ into its series form,}
=& 2\pi \int_0^\infty r^{2\alpha-1} \mathrm{e}^{- \frac{r^2}{\sigma^2}} \sum_{n=0}^\infty \frac{1}{n! \Gamma(n+1)}\left( \frac{|\mu|r}{\sigma^2} \right)^{2n} dr \\
=& 2\pi \sum_{n=0}^\infty \frac{|\mu|^{2n} \sigma^{-4n}}{n! \Gamma(n+1)} \int_0^\infty r^{2\alpha+2n-1} \mathrm{e}^{- \frac{r^2}{\sigma^2}} dr \\
\intertext{By the substitution $u = r^2$,}
=& 2\pi \sum_{n=0}^\infty \frac{|\mu|^{2n} \sigma^{-4n}}{n! \Gamma(n+1)} \int_0^\infty \sqrt{u}^{2\alpha+2n-1} \mathrm{e}^{- \frac{u}{\sigma^2}} \frac{1}{2 \sqrt{u}} du \\
=& \pi \sum_{n=0}^\infty \frac{|\mu|^{2n} \sigma^{-4n}}{n! \Gamma(n+1)} \int_0^\infty u^{(\alpha+n)-1} \mathrm{e}^{- \frac{u}{\sigma^2}} du \\
=& \pi \sum_{n=0}^\infty \frac{1}{n! \Gamma(n+1)} \left(\frac{|\mu|^2}{\sigma^4}\right)^{n} {(\sigma^2)}^{\alpha+n}\Gamma(\alpha+n) \\
\intertext{Using the Pochhammer symbol $(\alpha)_n$,}
=& \pi\sigma^{2\alpha} \Gamma(\alpha) \sum_{n=0}^\infty \frac{(\alpha)_n}{n!(1)_n} \left(\frac{|\mu|^2}{\sigma^2}\right)^{n} \\
=& \pi \sigma^{2\alpha} \Gamma(\alpha) {}_1F_1(\alpha;1;\frac{|\mu|^2}{\sigma^2}) \\
\intertext{Applying Kummer's transformation,}
=& \pi \sigma^{2\alpha} \Gamma(\alpha) \mathrm{e}^{\frac{|\mu|^2}{\sigma^2}} {}_1F_1(-\alpha+1;1;-\frac{|\mu|^2}{\sigma^2}) \\
=& \frac{\pi \sigma^{2\alpha} \Gamma(\alpha) L_{\alpha-1}(-\frac{|\mu|^2}{\sigma^2})}{ \mathrm{e}^{-\frac{|\mu|^2}{\sigma^2}} }.
\end{align}
This coincides with Eq.~\eqref{eq:a2}, and therefore
$\int_\mathcal{Z} p(z; \mu, \sigma^2, \alpha) dz = 1$.
\end{proof}

\section{Proof of $\sum_{n=0}^\infty p(n; \lambda, \alpha) = 1$}
\label{appendix:poiss_proof}

\begin{proof}
We show that the Poisson-type discrete distribution defined in Eq.~\eqref{eq:poiss_} satisfies the normalization condition: $\sum_{n=0}^\infty p(n; \lambda, \alpha) = 1$.
By direct manipulation of the series, we obtain
\begin{align}
\sum_{n=0}^\infty p(n; \lambda, \alpha)
&= \sum_{n=0}^\infty \frac{(\alpha)_n \lambda^n}{n!^2 \, L_{-\alpha}(\lambda)} \\
&= \sum_{n=0}^\infty \frac{(\alpha)_n \lambda^n}{n!^2 \, {}_1F_1(\alpha;1;\lambda)} \\
\intertext{Using the series expansion of the confluent hypergeometric function,}
&= \sum_{n=0}^\infty 
\frac{(\alpha)_n \lambda^n}
{n!^2 \, \displaystyle \sum_{n'=0}^\infty \frac{(\alpha)_{n'}}{n'!(1)_{n'}} \lambda^{n'}} \\
\intertext{Rewriting $(1)_{n'} = n'!$, we obtain}
&= \frac{\displaystyle \sum_{n=0}^\infty \frac{(\alpha)_n \lambda^n}{n!^2}}
{\displaystyle \sum_{n'=0}^\infty \frac{(\alpha)_{n'} \lambda^{n'}}{n'!^2}} \\
\intertext{Since the numerator and denominator are identical series,}
&= 1.
\end{align}
\end{proof}

\section{Computation of the acceptance probability $p$}
\label{appendix:acceptrate}

The acceptance probability $p$ in the Metropolis--Hastings algorithm described in Section~\ref{ssec:mh} is given by
\begin{align}
p &= \min(1,\, p'), \\
p' &= \frac{p(\tilde{n}'; \lambda,\alpha)}{p(\tilde{n}; \lambda,\alpha)}
\cdot \frac{q(\tilde{n})}{q(\tilde{n}')}.
\end{align}
Substituting the definitions of $p(\cdot)$ and $q(\cdot)$,
\begin{align}
p'
&= \frac{\lambda^{\tilde{n}'} (\alpha)_{\tilde{n}'}}{\tilde{n}'!^2 \, L(-\alpha,\lambda)}
\cdot \frac{\tilde{n}!^2 \, L(-\alpha,\lambda)}{\lambda^{\tilde{n}} (\alpha)_{\tilde{n}}}
\cdot \frac{\lambda^{\tilde{n}}}{\tilde{n}!}
\cdot \frac{\tilde{n}'!}{\lambda^{\tilde{n}'}} \\
\intertext{After cancellation of common factors,}
&= \frac{\tilde{n}! \, (\alpha)_{\tilde{n}'}}{\tilde{n}'! \, (\alpha)_{\tilde{n}}} \\
\intertext{Using $(\alpha)_n = \Gamma(\alpha+n)/\Gamma(\alpha)$,}
&= \frac{\tilde{n}! \, \frac{\Gamma(\alpha+\tilde{n}')}{\Gamma(\alpha)}}
{\tilde{n}'! \, \frac{\Gamma(\alpha+\tilde{n})}{\Gamma(\alpha)}} \\
&= \frac{\tilde{n}! \, \Gamma(\alpha+\tilde{n}')}{\tilde{n}'! \, \Gamma(\alpha+\tilde{n})}.
\end{align}

\section{Computation of the $n$th moment}
\label{appendix:nth-moment}

The $n$th moment in Eq.~\eqref{eq:mn} is computed as
\begin{align}
\mathbb{E}[x^n]
&= \int_0^\infty x^n p(x; \alpha, \beta, \lambda)\, dx \\
&= \frac{\beta^{\alpha}}{\Gamma(\alpha) L_{-\alpha}(\lambda)}
\int_0^\infty x^{n+\alpha-1} \exp(- \beta x)\,
I_0\!\left(2\sqrt{\beta\lambda x}\right) dx \\
\intertext{Using the integral identity involving the modified Bessel function,}
&= \frac{\beta^{\alpha}}{\Gamma(\alpha) L_{-\alpha}(\lambda)}
\cdot \frac{\Gamma(n+\alpha)\, L_{-n-\alpha}(\lambda)}{\beta^{n+\alpha}} \\
&= \frac{(\alpha)_n}{\beta^n}
\cdot \frac{L_{-\alpha-n}(\lambda)}{L_{-\alpha}(\lambda)}.
\end{align}





\bibliographystyle{unsrt}  
\bibliography{refs}






\end{document}